\documentclass[sigconf]{acmart}
\usepackage{algorithm}
\usepackage{algpseudocode}
\usepackage{listings}
\usepackage{multicol,multirow}
\usepackage{mathtools}
\usepackage{nomencl}
\usepackage{balance}

\AtBeginDocument{
  \providecommand\BibTeX{{%
    \normalfont B\kern-0.5em{\scshape i\kern-0.25em b}\kern-0.8em\TeX}}}

\begin{document}


\title{Using Decentralized Aggregation for Federated Learning with Differential Privacy}


\author{Hadeel Abd El-Kareem Abd El-Moaty Saleh}
\email{hadeelsaleh@aast.edu}
\additionalaffiliation{%
  \institution{College of Computing and Information Technology,
Arab Academy for Science, Technology, and Maritime Transport (AASTMT)}
  \city{Aswan}
  \country{Egypt}
   \postcode{81516}}
  
\author{Ana Fernández-Vilas}
\email{avilas@det.uvigo.es}
\orcid{0000-0003-1047-2143}
\additionalaffiliation{%
  \institution{atlanTTic - I \& C Lab - Universidade de Vigo}
  \city{Vigo}
  \country{Spain}
  \postcode{36310}}
  
 \author{Manuel Fernández-Veiga}
\email{mveiga@det.uvigo.es}
\orcid{0000-0002-5088-0881}
\additionalaffiliation{%
  \institution{atlanTTic - I \& C Lab - Universidade de Vigo}
  \city{Vigo}
  \country{Spain}
  \postcode{36310}
}
  
\author{Yasser El-Sonbaty}
\email{yasser@aast.edu}
\orcid{0000-0003-0140-1618}
\additionalaffiliation{%
 \institution{College of Computing and Information Technology,
Arab Academy for Science, Technology, and Maritime Transport (AASTMT)}
 \city{Alexandria}
 \country{Egypt}
  \postcode{1029}
}

\author{Nashwa El-Bendary}
\email{nashwa.elbendary@aast.edu}
\orcid{0000-0001-6553-4159}
\additionalaffiliation{%
  \institution{College of Computing and Information Technology,
Arab Academy for Science, Technology, and Maritime Transport (AASTMT)}
  \city{Aswan}
  \country{Egypt}
   \postcode{81516}
   }


\begin{abstract}
Nowadays, the ubiquitous usage of mobile devices and networks have raised concerns about the loss of control over personal data and research advance towards the trade-off between privacy and utility in scenarios that combine exchange communications, big databases and distributed and collaborative (P2P) Machine Learning techniques. On the other hand, although Federated Learning (FL) provides some level of privacy by retaining the data at the local node, which executes a local training to enrich a global model, this scenario is still susceptible to privacy breaches as membership inference attacks. To provide a stronger level of privacy, this research deploys an experimental environment for FL with Differential Privacy (DP) using benchmark datasets. The obtained results show that the election of parameters and techniques of DP is central in the aforementioned trade-off between privacy and utility by means of a classification example. 

\end{abstract}




\maketitle

\renewcommand{\shortauthors}{{Hadeel Abd El-Kareem Abd El-Moaty Saleh} et al.}

\section{Introduction}
\label{sec:intro}

Federated Learning (FL) offers a useful paradigm for training a Machine  Learning (ML) 
model from data distributed across multiple data  silos,  eliminating the need for raw 
data sharing as it has the ambition to  protect data privacy through distributed learning
methods that  keep the data local. In simple terms, with FL, it  is not the data that moves 
to a model, but it is a model that moves to  data, which means that training is happening 
from user interaction with end  devices. Federated Learning's key motivation is to provide  
privacy protection as well as there has recently been some  research into combining the 
formal privacy notion of  Differential  Privacy (DP) with FL. 

Instead of researching on the traditional FL model, we consider working on a peer-to-peer 
approach to FL or decentralized FL framework, where any node can play the role of model 
provider and model aggregator at different times. That is being applied depending on various 
aspects related to the communication infrastructure and the distribution of the data. For  
example, considering a traditional FL model where the aggregator is unavailable for some period. 
In a peer-to-peer approach, the  aggregator role can be played by a different node. Also, it can 
be decided from time to time that the aggregator node is the one with  more connections in the
network, or whatever other criteria that  allows the learning process to be improved. So, every 
node in the  network has the capability of being a model provider and an aggregator node, and it 
plays different roles in different rounds of  the learning process according to the state of the
infrastructure. However, under this scenario, where the communication infrastructure  cannot be 
considered reliable and secure, additional  privacy measures are required. In a decentralized FL 
system, while the data is kept local and not exchanged with peers, the model parameters could be 
observed by malicious or curious agents and be used to infer knowledge about the datasets or the 
aggregated model. To tackle this problem, a DP mechanism~\cite{X-jiang2021differential} can be 
introduced into the distributed learning environment, so that inference attacks are harder. 
However, as DP adds controlled noise to the information exchanged by the nodes, it might affect 
the learning rate and accuracy of the algorithm. This trade-off has been considered in some works 
for classical (centralized, single-server) FL systems~\cite{Z-zhao2021utility, Y-cao2020ifed}. 
In this paper, we study experimentally the interplay between DP and the learning performance in 
a distributed FL system.

The remaining of this paper is organized as follows. In Section~\ref{sec:related}, a review of 
the literature related to our work is presented. Section~\ref{sec:methods} starts with introducing
a background on DP and distributed FL methods, then the framework of the proposed system is 
described. The implementation details and experimental setup are illustrated in 
Section~\ref{sec:implementation}. The obtained numerical results are presented in 
Section~\ref{sec:results}. Section~\ref{sec:conclusions} discusses the attained observations, 
and provides several insights.

\section{Related Work}
\label{sec:related}

In the scientific literature, a number of studies have discussed on   distributed ML, FL, 
and privacy preserving machine learning. Reading diverse existing literature's  application, 
benefits, and drawbacks is one of the most significant   elements of study, we discovered that FL 
is a  relatively new topic with limited published literature and articles, but much research has 
already been conducted in the areas of privacy-preserving learning and distributed machine learning 
algorithms or the Traditional Centralized Federated Learning  which the role of the server goes to 
one role until convergence. 

In~\cite{1-zhang2021understanding}, the authors compared two different variants of  clipping 
for FedAvg: clip the client model vs. clip the client   model difference (CE-FedAvg), considering 
not all the clients participate in each round of communication. Here some conducted 
experiments for FedAvg, CE-FedAvg and DP-FedAvg with different models  namely MLP, AlexNet, Resnet, 
MobilNetV2, on two different benchmark  Dataset EMNIST and Cifar-10 for Classification, and local 
data distribution which falls into two ways; 1) IID Data setting where the samples are uniformly 
distributed to each client 2) Non-IID Data setting, where the clients have unbalanced samples. 
The results showed that the performance depends on the structures of the neural network being used 
and the heterogeneity data distribution among clients is one of the main causes of the different 
behavior between the clipped and unclipped.

The authors in~\cite{2-lin2020differential} proposed algorithms based on differentially private 
SGD (DP–SGD) that add Gaussian noises to each  computed gradient and then clip the noised gradient
(NC), which is  different to the conventional method in the sequence of clipping  gradient and 
adding noise (CA). The experimental settings to  verify the performance consider different 
factors, as the training of two popular deep learning models, (CNN and LSTM) on three  
different datasets, namely MNIST, CIFAR10, SVH, respectively, and  adoption of two gradient 
descent optimization methods (SGD and Adam) for evaluation, with the inclusion of
Gaussian noise and clipping. This work also proposes a new privacy protection metric called 
"Total  Parameters Value Difference" to measure the privacy protection  capability and examine 
how the impact of adding noise in the  training process will affect the model itself. The 
results showed  and validated the effectiveness of their proposed method (AC), which 
improves remarkably the accuracy of the  model when other  parameter settings are the same. 
However, the TPVD of AC are  lower than those of CA. The results showed proposed  a 
modification of AC, as showed changing the sequence of adding  noise and clipping can 
achieve higher accuracy and faster  convergence that outperforms the conventional method even  
under different parameter settings and the TPVD metric proposed  in this paper as a 
privacy protection metric for DL models can  better reflect the perturbation effects. 

In~\cite{3-choudhury2019differential}, the authors introduced a FL framework, with the 
proposed method of DL to add noise to the objective function of the optimization at each site 
to produce a minimizer of the perturbed objective. The proposed model is tested on two 
different datasets for two major tasks ;(1) prediction of adverse drug reaction (ADR), 
Dataset used Limited MarketScan Explorys Claims-EMR Data (LCED); (2) prediction of mortality 
rate, dataset using Medical Information Mart for Intensive Care (MIMIC III) data. The results 
show that although DL offers a strong level of privacy, it deteriorates the predictive capability 
of the produced global models due to the excessive amount of noise added during the distributed 
FL training process, while in~\cite{8-wei2020federated} the authors proposed a novel framework 
based on the concept of DP, in which artificial noises are added to the parameters 
at the clients side before aggregating, namely, noising before model aggregation FL (NbAFL), 
the proposed NbAFL evaluated by using multi-layer perception (MLP) on MNIST dataset, they 
found that there is an optimal K that achieves the best convergence performance at a fixed privacy 
level.

In~\cite{1-zhang2021understanding} the authors introduced a FL framework,
called collaborative FL (CFL), which enables edge devices to implement FL with
less reliance on a central controller to facilitate the deployment in IOT
applications. The concept of the proposed Framework based on this study where
some devices are directly connected to the Base station, while others are
associated with a certain number of neighboring devices. The main objective
was to overcome the challenge of energy limitations or a potentially high
transmission delay.  In order to overcome the short comings of the reviewed
literature, the main objective of the approach proposed in this paper is to
conduct an experimental framework to investigate the interaction between DP
and learning performance in a distributed FL system.

\section{Methods and Methodology}
 \label{sec:methods}
 
This section introduces the notion of Federated Learning considering guaranteeing the model 
privacy protection through utilizing the characteristic of keeping the data in the local  node.
Although FL keeps the data in the local  node, so that privacy is guaranteed, in order to protect 
the model. Also, we are going to apply DP mechanisms to the models which are exchanged among the 
nodes in the FL scenario. So, we are  also going to introduce the concept of DP. 

\subsection{Federated Learning: definition and role of  the aggregator node}

The notion of Federated Learning was first introduced in~\cite{4-mcmahan2017communication}, 
which demonstrates a new learning context in which a shared model is learned by aggregating 
locally computed gradient changes without centralizing different data on devices. 

Federated averaging (FedAvg) is a communication efficient  algorithm for distributed training 
with a number of clients. As it is mentioned in~\cite{5-zhang2021survey} its set-up is a system 
in which multiple  clients collaborate to solve machine learning problems, with a  central 
aggregator overseeing the process. This setting  decentralizes the training data, ensuring that 
each device's data is  secure. Federated learning is based on two key principles: local  computing 
and model transmission, which mitigates some of the  privacy risks and costs associated with 
standard centralized  machine learning methods. The client's original data is kept on site and 
cannot be transferred or traded. Each device uses local  data for local training, then uploads 
the model to the server for  aggregation, and finally, the server transmits the model update to  
the participants to achieve the learning goal. 

Formally~\cite{8-wei2020federated}, the server aggregates the weights sent from the $N$
clients as (FedAvg), as  
\begin{equation}
    \mathbf{w} = \sum_{i = 1}^N p_i \mathbf{w}_i
\end{equation}
where $\mathbf{w}_i$ is the parameter vector trained at the $i$-th client, $\mathbf{w}$ is the  
parameter vector after aggregating at the server, $N$ is the number of clients, and $p_i = 
|\mathcal{D}_i| / |\mathcal{D}|$, with $\mathcal{D}_i$ the dataset of node $i$ and $\mathcal{D} = 
\cup_i \mathcal{D}_i$ the whole distributed dataset. The server solves the optimization problem
\begin{equation}
    \mathbf{w}^\ast = \arg\min_{\mathbf{w}} \sum_{i = 1}^N p_i F_i(\mathbf{w}, \mathcal{D}_i)
\end{equation}
where $F_i$ is the local loss function of the $i$-th client. Generally, the local loss function 
is given by local empirical risks.

The training process of such a FL system usually contains the following four steps: 1) Local 
training: All active clients locally compute training gradients or parameters and send locally 
trained ML parameters to the server; 2) Model aggregating: The server performs secure aggregation 
over the uploaded parameters from N clients without learning local information; 3) Parameters
broadcasting: The server broadcasts the aggregated parameters to the N clients; 4) Model updating: 
All clients update their respective models with the aggregated parameters.

In order to prevent information leakage and the local model parameter which is circulated over 
the network from the inference attacks as it is vulnerable to it, a natural approach to defining
privacy for those models will be used namely Differential Privacy and this will be discussed in 
the next Section.

\subsection{Differential Privacy}

A Differential Privacy mechanism $M$ satisfies $(\epsilon, \delta)$-DP for two non-negative 
numbers $\epsilon$ and $\delta$, if the following inequality holds
\begin{equation}
    \mathbb{P}\bigl( M(\mathcal{D}) \in S \bigr) \leq \epsilon \mathbb{P} \bigl( M(\mathcal{D}^\prime) \in S \bigr) + \delta
\end{equation}
where $\mathcal{D}$ and $\mathcal{D}^\prime$ are neighboring datasets under the Hamming  
distance, and $S$ is an arbitrary subset of outputs of $M$.

Intuitively speaking, the number $\delta$ represents the probability that a  mechanism’s output 
varies by more than a factor $\epsilon$ when  applied to a dataset and any one of its close 
neighbors. A lower value of $\delta$ signifies greater confidence and a smaller value of $\epsilon$ 
tightens the  standard for privacy protection \cite{6-dwork2014algorithmic, 7-abadi2016deep}. 
Typical mechanisms for $M$ include the perturbation of the dataset values with Laplacian, 
Exponential, or Gaussian noise. 

In order for the perturbation mechanisms to have formal privacy  guarantees, the amount of noise 
that is added to the local model  updates across each provider may result in exploding gradients  
problem, which refers to large increases in the norm of the  gradient during training, so it 
requires a clipping operation, and  this will be discussed in the next Section. 

\subsection{Clipping and Bounded Norm Operation}

As discussed in~\cite{7-abadi2016deep}, clipping is a crucial step in ensuring the DP  of FL 
algorithms. so, each provider’s/client’s model update needs  have a bounded norm, which is 
ensured by applying an operation  that shrinks individual model updates when their norm exceeds 
a  given threshold. To create FL algorithms that protect DP is to  know that clipping impacts a 
FL algorithm's convergence  performance. There are two major clipping strategies used for FL 
algorithms, one is local model clipping which consists in the clients directly  clipping the 
models sent to the server; the other is difference  clipping, where the local update difference
between the initial  model and the output model i clipped. 

\subsection{Proposed Framework of P2P FL with DP}

In this Section we will define the peer-to-peer framework that  worked on and how the DP will 
be introduced in some of the nodes and finally assess the results by measuring  
some metrics. 

\begin{figure*}
    \centering
    \includegraphics[scale=0.5]{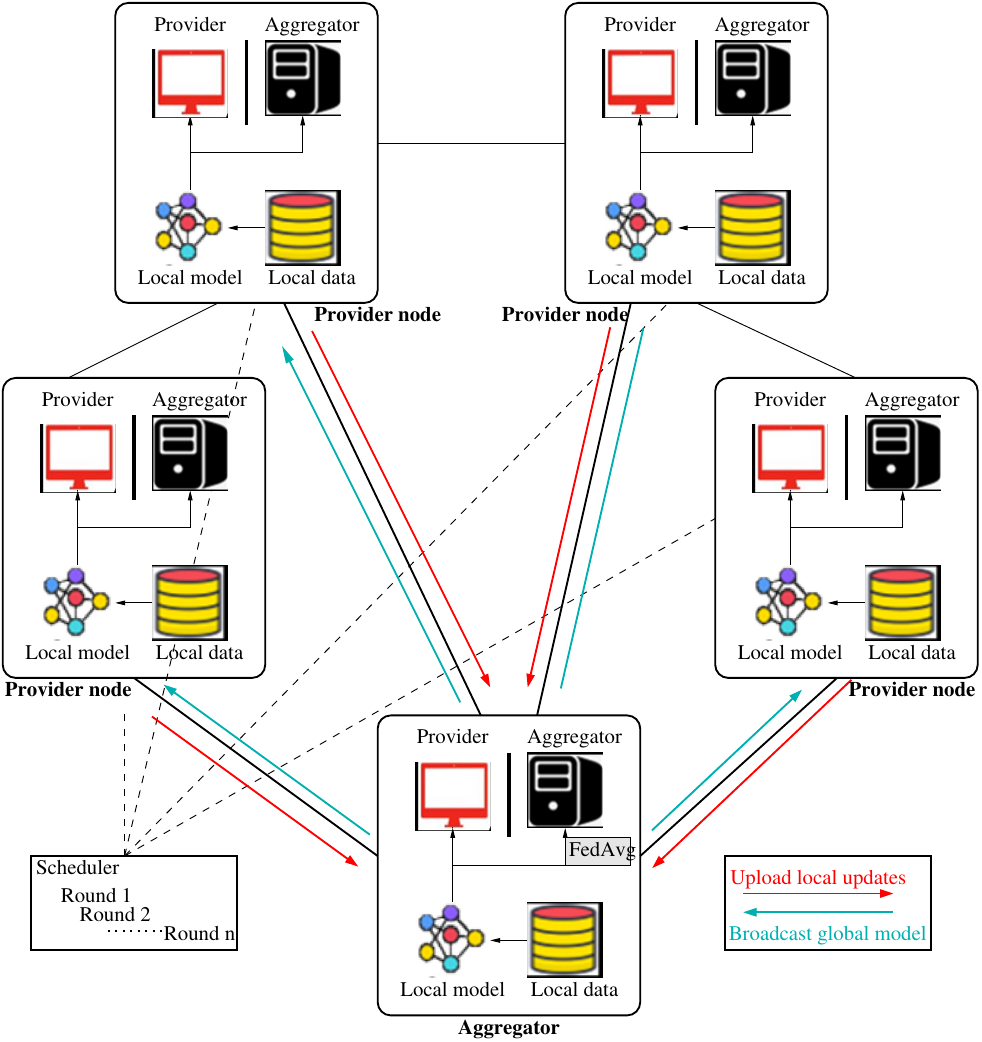}
    \caption{P2P FL Architecture}
    \label{fig:model}
\end{figure*}
We consider a FL problem in a decentralized setting as shown in  Fig.~\ref{fig:model}, in which 
a set $V = \{1, \dots, K \}$ of nodes can only communicate with their respective neighbors. 
Each node $i \in V$ stores a subset $\mathcal{D}_i$ of samples of a common unknown 
distribution $\mathcal{D}$ from which  we are interested in learning. The connectivity 
is characterized by an undirected graph $G = (V, E)$, with $V$ denoting the set of nodes and 
$E \subseteq \{(i, j) \in V \times V: i \neq j \}$ the set of edges. The set of neighbors of 
node $i$ is denoted as $N_i = \{ j \in V:  (i, j) \in E \}$.

Each node has available a local data set $\mathcal{D}_i$, and all devices collaboratively 
train a machine learning model by exchanging  model-related information without directly 
disclosing  data samples to one another. Each node can play the role of aggregator  of models 
or provider of a model. 
\begin{itemize}
    \item When a node $N$ is an aggregator, it receives the model  from all its neighbors and 
    obtains an aggregated  model. 
    \item When a node $N$ is a provider, it sends its model  parameter to the aggregator/aggregators. 
\end{itemize}

The Peer to Peer FL scenario works in rounds. A round is defined  in the graph by identifying 
the node which is the aggregator. The  rounding process until the learning process converges. 
A round is defined in the graph by identifying the node which is the  aggregator. The aggregator
broadcasts the global model Parameter  to its neighbor providers. Then sampled clients perform 
local training of (e.g., SGD optimizer) and compute their updates.

\begin{figure}
    \centering
    \includegraphics[scale=0.75]{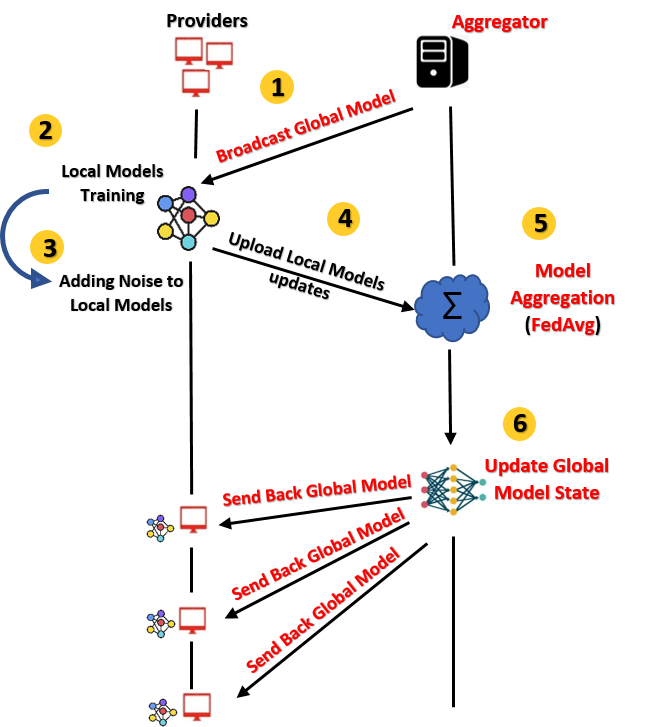}
    \caption{Communication in P2P FL}
    \label{fig:FL-communication}
\end{figure}

To introduce Differential Privacy in the previous P2P FL model, a new graph is considered as 
follows. An undirected differential  private graph DPG $(N, E, \mathsf{DP})$ with $\mathsf{DP}$ 
denoting the applying  DP for a node $i$, assuming that $\mathsf{DP}(i) \in \{0, 1 \}$ represents 
whether the node applies differential  Privacy when obtaining/updating the local Model. Also, 
the  rounding process is updated, so that the perturbation mechanisms  (DP) are added to the 
local updates for a subset of nodes. In order  for the perturbation mechanism to have formal 
privacy  guarantees, each local update needs to have a bounded norm,  which is ensured by applying 
a clipping operation that reduces the  clients’ individual model updates when their norm exceeds a
given threshold. The network and communication model are depicted in Figures~\ref{fig:model}
and~\ref{fig:FL-communication}, respectively.

Measuring quality of DP P2P FL: mainly, we will focused on the relationship between the privacy 
budget and the utility of the  model as well as the performance of the attacker so, we are  
interested in studying the trade-off utility (accuracy, loss ) vs.  privacy (epsilon, strength 
against model attacks) in them: (1) Privacy: evaluation of how much information is leaked by the  
DF mechanisms; and (2) Utility: evaluation of the difference  between results obtained from the
original and deferentially  private data (Loss - Accuracy) 

\section{Implementation Details and Experimental Setup}
\label{sec:implementation}

\begin{algorithm}[t]
  \caption{\label{alg:scheduler} Computations in the peer-to-peer FL model.}
  \begin{algorithmic}[1]
    \State \textbf{Initialize}: for all $i = 1, \dots, K$, $X_i^{(0)} = x^{(0)}$ 
    \For {each round $t = 0, 1, \dots$}
    \For {each node $i = 1, \dots, K$ do in parallel}
    \State Update nodes $x_i^{(t, 0)} = x^{(t)}$
        \For {each neighbor node $j \in N_i$ do in parallel}
        \State \textbf{local update}: compute stochastic gradient and clipping
        \State{DP}: introduce noise into the model update
        \State \textbf{Local upload}: send model to aggregator node
        \EndFor
    \State \textbf{Global averaging}: average the model parameters at the server
    \EndFor
    \EndFor
  \end{algorithmic}
  \end{algorithm}

For system implementation, framing the experiments in previously described way makes them 
manageable by  Anaconda, which is a Python-based data processing platform that  runs on 
different environment management systems like  Windows, macOS and Linux. It can easily 
create, save, load, and  switch between environments on the local computer as it comes  with 
some default implementations of Integrated Development  Environments (IDEs), the one used 
is Spyder IDE, which is an editor with syntax highlighting, introspection, and code completion. 

Also, we used a set of frameworks, libraries, and dependencies to  meet all the requirements 
in a smooth environment for our peer-to-peer framework. 

The \texttt{deepee} differentially private PyTorch library is an open source deep learning 
framework for developing deep learning models with the Opacus library that enables training 
PyTorch models with differential  privacy. Also, it supports training with minimal code changes 
required on the client and allows tracking the privacy budget expended at any given moment. 

Also, one of the most important Framework used is \texttt{Flwr} for building FL systems. It is
used for edge devices to collaboratively learn a shared prediction model, while keeping  their 
training data on the device. 

Algorithm~\ref{alg:scheduler} describes aggregator execution pseudocode for Federated Averaging
targeting updates from $K$ providers per round in peer-to-peer.

\label{sec:setup}

\begin{table*}[t]
    \centering
    \small
    \begin{tabular}{cl|cccccccccc}
        & & \multicolumn{2}{c}{\textsc{client 1}}
        & \multicolumn{2}{c}{\textsc{client 2}}
        & \multicolumn{2}{c}{\textsc{client 3}}
        & \multicolumn{2}{c}{\textsc{client 4}}
        & \multicolumn{2}{c}{\textsc{client 5}} \\
        \textsc{round} & & MNIST & CIFAR & MNIST & CIFAR & MNIST & CIFAR & MNIST & CIFAR & MNIST & CIFAR \\ \hline
        \multirow{2}{*}{1} & Loss & $0.125$ & $2.287$ & $0.131$ & $2.286$ & $0.124$ & $2.287$ & $0.129$ & $2.283$ & $0.116$ & $2.278$ \\
        & Accuracy & $0.959$ & $0.151$ & $0.964$ & $0.150$ & $0.958$ & $0.144$ & $0.958$ & $0.155$ & $0.964$ & $0.151$ \\ \hline
        \multirow{2}{*}{2} & Loss & $0.081$ & $1.351$ & $0.085$ & $1.333$ & $0.068$ & $1.360$ & $0.076$ & $1.326$ & $0.075$ & $1.370$ \\
        & Accuracy & $0.975$ & $0.548$ & $0.971$ & $0.562$ & $0.977$ & $0.545$ & $0.973$ & $0.547$ & $0.977$ & $0.538$ \\ \hline
        
        \multirow{2}{*}{3} & Loss & $0.078$ & $0.982$ & $0.079$ & $0.965$ & $0.064$ & $0.979$ & $0.071$ & $0.970$ & $0.074$ & $0.969$ \\
        & Accuracy & $0.978$ & $0.667$ & $0.974$ & $0.674$ & $0.979$ & $0.666$ & $0.979$ & $0.676$ & $0.978$ & $0.660$\\ \hline
        
        \multirow{2}{*}{4} & Loss & $0.078$ & $0.952$ & $0.075$ & $0.899$ & $0.061$ & $0.9902$ & $0.068$ & $0.943$ & $0.07$ & $0.879$ \\
        & Accuracy & $0.979$ & $0.693$ & $0.975$ & $0.715$ & $0.979$ & $0.712$ & $0.978$ & $0.705$ & $0.978$ & $0.705$ \\ \hline
        
        \multirow{2}{*}{5} & Loss & $0.078$ & $0.979$ & $0.072$ & $0.958$ & $0.060$ & $0.983$ & $0.065$ & $0.993$ & $0.072$ & $0.985$ \\
        & Accuracy & $0.979$ & $0.714$ & $0.976$ & $0.731$ & $0.980$ & $0.720$ & $0.979$ & $0.710$ & $0.979$ & $0.727$ \\ \hline
        \end{tabular}
        \caption{Performance for standard FL without DP. Clients}
    \label{tab:table1}
\end{table*} 

\begin{table*}[t]
    \centering
    \small
    \begin{tabular}{cl|cccccccccc}
        & & \multicolumn{2}{c}{\textsc{client 1}}
        & \multicolumn{2}{c}{\textsc{client 2}}
        & \multicolumn{2}{c}{\textsc{client 3}}
        & \multicolumn{2}{c}{\textsc{client 4}}
        & \multicolumn{2}{c}{\textsc{client 5}} \\
        \textsc{round} & & MNIST & CIFAR & MNIST & CIFAR & MNIST & CIFAR & MNIST & CIFAR & MNIST & CIFAR \\ \hline
        \multirow{2}{*}{1} & Loss & $0.093113$ & $0.073505$ & $0.089048$ & $0.0735248$ & $0.087845$ & $0.0735352$ & $0.0903270$ & $0l.0735159$ & $0.0931194$ & $0.0734737$ \\
        & Accuracy & $0.695$ & $0.163$ & $0.708333$ & $0.147$ & $0.714186$ & $0.132$ & $0.70333$ & $0.14$ & $0.6975$ & $0.177$ \\\hline
         \multirow{2}{*}{2} & Loss &  $0.050274$ & $0.070985$ & $0.044131$ & $0.0710392$ & $0.046384$ & $0.0711034$ & $0.0497613$ & $0.0710027$ & $0.0532784$ & $0.0708465$ \\
         & Accuracy & $0.86$ & $0.187$ & $0.87166$ & $0.186$ & $0.87166$ & $0.3203$ & $0.85083$ & $0.187$ & $0.8525$ & $0.199$ \\ \hline 
         \multirow{2}{*}{3} & Loss & $0.047559$ & $0.066884$ & $0.041380$ & $0.0669977$ & $0.044854$ & $0.0670923$ & $0.0497825$ & $0.0669886$ & $0.0540392$ & $0.0667669$ \\
         & Accuracy & $0.885$ & $0.229$ & $0.894166$ & $0.221$ & $0.8866$ & $0.234$ & $0.869186$ & $0.198$ & $0.875$ & $0.259$ \\ \hline
         \multirow{2}{*}{4} & Loss & $0.048086$ & $0.065885$ & $0.039818$ & $0.0656856$ & $0.043185$ & $0.0655003$ & $0.0488195$ & $0.0658598$ & $0.0534120$ the & $0.0654942$ \\
         & Accuracy & $0.905$ & $0.256$ & $0.9033$ & $0.247$ & $0.90583$ & $0.258$ & $0.8841$ & $0.227$ & $0.8983$ & $0.258$ 
         \\ \hline
         \multirow{2}{*}{5} & Loss & $0.04594$ & $0.065530$ & $0.037074$ & $0.0651977$ & $0.041597$ & $0.0648178$ & $0.0448278$ & $0.0655778$ & $0.04965$ & $0.06506$ \\
         & Accuracy & $0.90583$ & $0.275$ & $0.9116$ & $0.258$ & $0.908$ & $0.263$ & $0.9001$ & $0.243$ & $0.9025$ & $0.28$ \\ \hline
    \end{tabular}
    \caption{Performance for standard FL with DP. Clients}
    \label{tab:table2}
\end{table*}

\begin{table}[hpb]
    \centering
    \begin{tabular}{ccccc}
    & \multicolumn{2}{c}{MNIST} & \multicolumn{2}{c}{CIFAR} \\ 
    & \textsc{accuracy} & \textsc{loss} & \textsc{accuracy} & \textsc{loss} \\ \hline
    1 & $0.70366658$ & $0.09069073$ &  $0.3114$ & $1.853063607$ \\
    2 & $0.8613333$ & $0.048766047$ & $0.4782$ & $1.146727948$ \\
    3 & $0.88199$ &	$0.047523097$ & $0.5142$ & $1.373809791$ \\
    4 & $0.899933$ & $0.046664422$ & $0.5128$ & $1.364574289$ \\
    5 & $0.90599$ &	$0.043820837$ & $0.5374$ & $1.333600569$ \\ \hline
    \end{tabular}
    \caption{Performance for standard FL w/o DP. Server}
    \label{tab:table3}
\end{table}

\begin{table}[hpb]
    \centering
    \begin{tabular}{ccccc}
    & \multicolumn{2}{c}{MNIST} & \multicolumn{2}{c}{CIFAR} \\ 
    & \textsc{accuracy} & \textsc{loss} & \textsc{accuracy} & \textsc{loss} \\ \hline
    1 & $0.70366658$ & $0.09069073$ & $0.1518$ & $0.0735108$ \\
    2 & $0.8613333$ & $0.048766047$ & $0.15586$ & $0.071215$ \\
    3 & $0.88199$ & $0.047523097$ & $0.2282$ & $0.0669454$ \\
    4 & $0.899933$ & $0.046664422$ & $0.2492$ & $0.0656846$ \\
    5 &  $0.90599$ & $0.043820837$ & $0.2638$ & $0.0652362$\\ \hline
    \end{tabular}
    \caption{Performance for standard FL with DP. Server}
    \label{tab:table4}
\end{table}

\begin{table*}[t]
    \centering
    \small
    \begin{tabular}{cl|cccccccccc}
        & & \multicolumn{2}{c}{\textsc{client 1}}
        & \multicolumn{2}{c}{\textsc{client 2}}
        & \multicolumn{2}{c}{\textsc{client 3}}
        & \multicolumn{2}{c}{\textsc{client 4}}
        & \multicolumn{2}{c}{\textsc{client 5}} \\
        \textsc{round} & & MNIST & CIFAR & MNIST & CIFAR & MNIST & CIFAR & MNIST & CIFAR & MNIST & CIFAR \\ \hline
        \multirow{2}{*}{1} & Loss &  $0.070$ & $1.874$ & \multirow{2}{*}{$\mathbf{0.085013 *}$} & \multirow{2}{*}{$\mathbf{1.854197* }$} & $0.079$ & $1.837$ & $0.108$ & $1.886$ & $0.083$ & $1.840$ \\
        & Accuracy &  $0.977$ & $0.350$ & & & $0.975$ & $0.360$ & $0.973$ & $0.3444$ & $0.96$ & $0.346$ \\ \hline
        
        \multirow{2}{*}{2} & Loss & \multicolumn{2}{c}{\multirow{2}{*}{\textcolor{red}{DNP}}} & $0.163$ & $1.864$ & $0.149$ & $1.844$ & \multirow{2}{*}{$\mathbf{0.15307}$ *} & \multirow{2}{*}{$\mathbf{1.85297}$ *} & $0.147$ & $1.851$ \\
        & Accuracy & & & $0.947$ & $0.323$ & $0.955$ & $0.329$ & & & $0.958$ & $0.340$ \\ \hline
        
        \multirow{2}{*}{3} & Loss & $0.137$ & $1.834$ & $0.136$ & $1.809$ & \multicolumn{2}{c}{\multirow{2}{*}{\textcolor{red}{DNP}}} & $0.133$ & $1.839$ & \multirow{2}{*}{$\mathbf{0.13665 *}$} & \multirow{2}{*}{$\mathbf{1.8271}$ *} \\
        & Accuracy & $0.951$ & $0.311$ & $0.955$ & $0.323$ & & & $0.952$ & $0.320$ & & \\ \hline
        
        \multirow{2}{*}{4} & Loss & \multirow{2}{*}{$\mathbf{0.20864 *}$} & \multirow{2}{*}{$\mathbf{1.71693 *}$} & $0.227$ & $1.719$ & \multicolumn{2}{c}{\multirow{2}{*}{\textcolor{red}{DNP}}} & \multicolumn{2}{c}{\multirow{2}{*}{\textcolor{red}{DNP}}} & $0.190$ & $1.715$ \\
        & Accuracy & & & $0.925$ & $0.393$ & & & & & $0.930$ & $0.396$ \\ \hline
        
        \multirow{2}{*}{5} & Loss & \multicolumn{2}{c}{\multirow{2}{*}{\textcolor{red}{DNP}}} & $0.220$ & $1.625$ & \multirow{2}{*}{$\mathbf{0.2226 *}$} & \multirow{2}{*}{$\mathbf{1.63389 *}$} & $0.226$ & $1.643$ & \multicolumn{2}{c}{\multirow{2}{*}{\textcolor{red}{DNP}}} \\
        & Accuracy & & & $0.923$ & $0.421$ & & & $0.924$ & $0.415$ & & \\ \hline
    \end{tabular}
    \caption{Performance for decentralized FL w/o DP. (DNP: Does Not Participate); (* the node is the aggregator in this round)}
    \label{tab:table5}
\end{table*}    

\begin{table*}[t]
    \centering
    \small
    \begin{tabular}{cl|cccccccccc}
        & & \multicolumn{2}{c}{\textsc{client 1}}
        & \multicolumn{2}{c}{\textsc{client 2}}
        & \multicolumn{2}{c}{\textsc{client 3}}
        & \multicolumn{2}{c}{\textsc{client 4}}
        & \multicolumn{2}{c}{\textsc{client 5}} \\
        \textsc{round} & & MNIST & CIFAR & MNIST & CIFAR & MNIST & CIFAR & MNIST & CIFAR & MNIST & CIFAR \\ \hline
        
        \multirow{2}{*}{1} & Loss &  $0.1434536$ & $0.0735429$ & \multirow{2}{*}{$\mathbf{0.1421104 *}$} & \multirow{2}{*}{$\mathbf{0.073515 *}$} & $0.1442883$ & $0.073528$ & $0.140833$ & $0.073529$ & $0.142315$ & $0.073507$ \\
        & Accuracy & $0.50008$ & $0.178$ & & & $0.49333$ & $0.14$ & $0.515$ & $0.161$ & $0.5291$ & $0.172$ \\ \hline
        
        \multirow{2}{*}{2} & Loss & \multicolumn{2}{c}{\multirow{2}{*}{\textcolor{red}{DNP}}} & $0.04892$ & $0.07189$ & $0.05340$ & $0.07190$ & \multirow{2}{*}{$\mathbf{0.052364 *}$} & \multirow{2}{*}{$\mathbf{0.071975} *$} & $0.054835$ & $0.07194$ \\
        & Accuracy & & & $0.8383$ & $0.209$ & $0.9266$ & $0.201$ & & & $0.835$ & $0.186$ \\ \hline 
        
        \multirow{2}{*}{3} & Loss & $0.041484$ & $0.067329$ & $0.03560$ & $0.06895$ & \multicolumn{2}{c}{\multirow{2}{*}{\textcolor{red}{DNP}}} & $0.04003$ & $0.068072$ & \multirow{2}{*}{$\mathbf{0.040854 *}$} & \multirow{2}{*}{$\mathbf{0.0682872 *}$} \\
        & Accuracy & $0.885$ & $0.268$ & $0.8883$ & $0.256$ & & & $0.8808$ & $0.228$ & & \\ \hline
        
        \multirow{2}{*}{4} & Loss & \multirow{2}{*}{$\mathbf{0.42075 *}$} & \multirow{2}{*}{$\mathbf{0.06580 *}$} & $0.03644$ & $0.066814$ & \multicolumn{2}{c}{\multirow{2}{*}{\textcolor{red}{DNP}}} & 
        \multicolumn{2}{c}{\multirow{2}{*}{\textcolor{red}{DNP}}} & $0.04480$ & $0.06698$ \\
        & Accuracy & & & $0.89916$ & $0.284$ & & & & & $0.90083$ & $0.26$ \\ \hline
        
        \multirow{2}{*}{5} & Loss & \multicolumn{2}{c}{\multirow{2}{*}{\textcolor{red}{DNP}}} & $0.03356$ & $0.06569$ & \multirow{2}{*}{$\mathbf{0.0402994 *}$} & \multirow{2}{*}{$\mathbf{0.0650621 *}$} & $0.038218$ & $0.0658201$ & \multicolumn{2}{c}{\multirow{2}{*}{\textcolor{red}{DNP}}} \\
        & Accuracy & & & $0.91166$ & $0.298$ & & & $0.90666$ & $0.258$ & & \\ \hline 
    \end{tabular}
    \caption{Performance for decentralized FL with DP. (DNP: Does Not Participate); (* the node is the aggregator in this round)} 
    \label{tab:table6}
\end{table*}

The main goal of the experiment is to know whether DL will affect the Learning process as we
mentioned before in the problem as well as to get over the privacy concerns and to 
privacy-preserving guarantee.

We consider working on two benchmarks dataset MNIST and CIFAR-10. MNIST has been used in many 
research experiments. It is a large database of handwritten digits of black and white images 
from NIST's original datasets which is a large database of handwritten uppercase and lowercase 
letters as well as digits. It is normalized to fit into a 28x28 pixel, which introduces grayscale
levels, the dataset included images only of handwritten digits database contains 60,000 training 
images and 10,000 testing images in which, half of the training set and half of the test set 
were taken from NIST's training dataset, while the other half of the training set and the 
other half of the test set were taken from MNIST's testing dataset. While the other one is 
CIFAR-10 which consists of $60000$ $32 \times 32$ color images in 10 classes, with $6000$ 
images per class. There are 50000 training images and 10000 test images, the dataset is divided 
into five training batches and one test batch, each with $10000$ images. The test batch contains
exactly $1000$ randomly selected images from each class. The training batches contain the 
remaining images in random order but some training batches may contain more images from one  
class than another. Between them, the training batches contain  exactly 5000 images from each 
class. The CIFAR-10 classes are  airplane, automobile, bird, cat, deer, dog, frog, horse, ship, 
and truck. 

We are going to propose a set of successive Experiments confined to a FL problem in a 
decentralized setting with applied DL mechanisms to some nodes depending on some aspects 
related to the state of the infrastructure of our Peer-to-Peer Framework.

The entire experiments will be having:
\begin{itemize}
\item a scheduler to regulate (control) the process of the Peer-to-Peer Network in each round 
for all the nodes as to set a role for each node in the Network to be an aggregator from the 
neighbor nodes or the provider.

\item Also, the scheduler will control whether the node has DL or not, depending on two aspects:
\begin{itemize}
\item The number of samples in dataset (size of dataset).
\item The number of connections with other nodes.
\end{itemize}
\end{itemize}
Here, the experiments run on 5 clients with equalized data but it is split randomly.

For the setup of peer to peer, the role of the aggregator goes to the node which has the highest 
number of connections with the other nodes while its neighbors play the role of provider as well 
as all of them will have DL and so on in each successive round.

\section{Experimental Results and Discussion}
\label{sec:results}

This Section shows the extensive evaluations were conducted to show the performance of  the 
proposed methods as there are the  results of four different approaches  Centralized (Traditional) 
Federated  Learning, Differential Privacy to centralized Federated Learning, peer-to-peer 
Federated Learning and Differential Privacy to Peer to Peer on two different Datasets MNIST 
and CIFAR in Five Rounds.

The experiments show when it runs on five clients with equalized data among all the clients with 
a single aggregator (server) until convergence achieved, and this is go to the part of the 
traditional FL. On the other hand, the second part of the experiments is when it runs on peer 
to peer FL, the rule of the aggregator does not go to a specific node among all the rounds but 
it depends on the node with the highest connections among the nodes and it differs from round to 
round in which every node can play either one rule of aggregator or provider per round until the
learning process over all the rounds finishes. 

Tables~\ref{tab:table1} and~\ref{tab:table2} list the baseline results for the two datasets in 
a classical FL setting, namely with a fixed node acting as a server for $5$ clients, wherein 
the dataset is uniformly split among the clients. The final accuracy and loss attained for 
standard FL without DP and with DP Server, respectively, are listed in Tables~\ref{tab:table3} 
and~\ref{tab:table4}, also for both datasets.

\begin{table}[t]
    \centering
    \begin{tabular}{ccccc}
    & \multicolumn{2}{c}{MNIST} & \multicolumn{2}{c}{CIFAR} \\ 
    & \textsc{accuracy} & \textsc{loss} & \textsc{accuracy} & \textsc{loss} \\ \hline
    1 & $0.97525$ & $0.0850137$ & $0.35$ & $1.854197413$ \\
    2 & $0.96233$ & $0.15307427$ & $0.3307$ & $1.85297886$ \\
    3 & $0.95267$ & $0.136656276$	 & $0.318$ & $1.827163457$ \\
    4 & $0.9275$ & $0.20864009$ & $0.3945$ & $1.71693456$ \\
    5 & $0.9235$ & $0.222623288$ & $0.418$ & $1.633897543$ \\ \hline
    \end{tabular}
    \caption{Performance for decentralized FL w/o DP: aggregator nodes.}
    \label{tab:table7}
\end{table}
\begin{table}[t]
    \centering
    \begin{tabular}{ccccc}
    & \multicolumn{2}{c}{MNIST} & \multicolumn{2}{c}{CIFAR} \\ 
    & \textsc{accuracy} & \textsc{loss} & \textsc{accuracy} & \textsc{loss} \\ \hline
    1 & $0.50773$ & $0.142110457$ & $0.16275$ & $0.073515633$ \\
    2 & $0.83334$ & $0.052364741$ & $0.195$ & $0.071975463$ \\
    3 & $0.88471$ & $0.040853945$ & $0.248$ & $0.068287218$ \\
    4 & $0.9$ & $0.042075122$ & $0.272$  & $0.06580271$ \\
    5 & $0.9092$ & $0.040299454$ & $0.278$ & $0.065062102$ \\ \hline
    \end{tabular}
    \caption{Performance for decentralized FL with DP: aggregator nodes.}
    \label{tab:table8}
\end{table}

In turn, Tables~\ref{tab:table5} and~\ref{tab:table6} list the 
performance (loss and accuracy) for the case of decentralized FL, 
where the aggregator is changing in each round (the aggregator node 
is marked in boldface numbers). The nodes that do not participate 
in a given round are due to the graph connectivity, recall that our
graph is not complete, so in a particular round there might be no 
physical communication between a client and the aggregator node. 
Even in absence of DP, this could slow down the rate of learning 
for the whole system in comparison to a full interchange in a round,
as in classical FL.

According to the results, when comparing the centralized FL and DP centralised FL:  
accuracy on MNIST is consistently high for all the rounds, so little sensitive to the 
introduction of privacy; in contrast, accuracy on CIFAR is initially low and gradually 
increases up to $70$\% in round $5$. As expected, Table~\ref{tab:table7} shows that, in 
standard FL, with and without DP at the clients, performance is homogeneous across all the 
clients, which attain similar performance. This is simply an indication that the data have been 
split uniformly without bias among the clients. Note also that, when DP is introduced 
(Table~\ref{tab:table8}, performance decreases drastically for CIFAR, while is stable for MNIST.
Thus, we clearly see that DP is more effective (for a constant $\epsilon$) in the latter case, 
and that the tuning of DP needs to be carefully set depending on the statistical distribution of 
the data. We also confirm experimentally that DP has a substantial impact on accuracy, and can
therefore make convergence quite slow in FL.

\section{Conclusions}
\label{sec:conclusions}

This paper has analyzed a decentralized approach for a Federated  Learning, namely peer-to-peer
FL with Differential Privacy to enable privacy among the network nodes and significantly improve 
upon established ways for protecting  privacy. Peer-to-peer FL  with DP (with noise multiplier 
$0.5$, $\epsilon = 2.59$ for every client) is able to protect personal data much better than 
the traditional methods. We have compared as well different approaches to ML according to privacy: 
1) Centralized FL (reduction of the P2P model where the aggregator is the same in all rounds); 2) 
P2P FL (decentralized); 3) DP Centralized ; 4) DP P2P FL. 

Strong privacy requirements in FL among a fully decentralized set of agents can be achieved by 
adapting the natural solution of introducing DL at the clients in each communication and 
computing round. However, in contrast to the well-studied case of a FL approach with DP in the
scenarios of a single and central aggregator, the peer-to-peer interactions during all the 
learning cycle (including local learning, noise injection and distribution of the models) is 
more complex and still not understood. In this respect, the work in this paper are a first 
experimental attempt at getting some insight into this interplay FL vs. DP vs. communications. 


\begin{acks}
  This work was supported by the Spanish Government under research project ``Enhancing Communication Protocols with Machine Learning while Protecting Sensitive Data (COMPROMISE)" PID2020-113795RB-C33, funded by MCIN/AEI/10.13039/501100011033
\end{acks}


\bibliographystyle{ACM-Reference-Format}
\balance
\bibliography{references}


\begin{thebibliography}{11}


\ifx \showCODEN    \undefined \def \showCODEN     #1{\unskip}     \fi
\ifx \showDOI      \undefined \def \showDOI       #1{#1}\fi
\ifx \showISBNx    \undefined \def \showISBNx     #1{\unskip}     \fi
\ifx \showISBNxiii \undefined \def \showISBNxiii  #1{\unskip}     \fi
\ifx \showISSN     \undefined \def \showISSN      #1{\unskip}     \fi
\ifx \showLCCN     \undefined \def \showLCCN      #1{\unskip}     \fi
\ifx \shownote     \undefined \def \shownote      #1{#1}          \fi
\ifx \showarticletitle \undefined \def \showarticletitle #1{#1}   \fi
\ifx \showURL      \undefined \def \showURL       {\relax}        \fi
\providecommand\bibfield[2]{#2}
\providecommand\bibinfo[2]{#2}
\providecommand\natexlab[1]{#1}
\providecommand\showeprint[2][]{arXiv:#2}

\bibitem[Abadi et~al\mbox{.}(2016)]%
        {7-abadi2016deep}
\bibfield{author}{\bibinfo{person}{Martin Abadi}, \bibinfo{person}{Andy Chu}, \bibinfo{person}{Ian Goodfellow}, \bibinfo{person}{H~Brendan McMahan}, \bibinfo{person}{Ilya Mironov}, \bibinfo{person}{Kunal Talwar}, {and} \bibinfo{person}{Li Zhang}.} \bibinfo{year}{2016}\natexlab{}.
\newblock \showarticletitle{Deep learning with differential privacy}. In \bibinfo{booktitle}{\emph{Proceedings of the 2016 ACM SIGSAC conference on computer and communications security}}. \bibinfo{pages}{308--318}.
\newblock


\bibitem[Cao et~al\mbox{.}(2020)]%
        {Y-cao2020ifed}
\bibfield{author}{\bibinfo{person}{Hui Cao}, \bibinfo{person}{Shubo Liu}, \bibinfo{person}{Renfang Zhao}, {and} \bibinfo{person}{Xingxing Xiong}.} \bibinfo{year}{2020}\natexlab{}.
\newblock \showarticletitle{IFed: A novel federated learning framework for local differential privacy in Power Internet of Things}.
\newblock \bibinfo{journal}{\emph{International J. of Distributed Sensor Networks}} \bibinfo{volume}{16}, \bibinfo{number}{5} (\bibinfo{year}{2020}), \bibinfo{pages}{1550147720919698}.
\newblock


\bibitem[Choudhury et~al\mbox{.}(2019)]%
        {3-choudhury2019differential}
\bibfield{author}{\bibinfo{person}{Olivia Choudhury}, \bibinfo{person}{Aris Gkoulalas-Divanis}, \bibinfo{person}{Theodoros Salonidis}, \bibinfo{person}{Issa Sylla}, \bibinfo{person}{Yoonyoung Park}, \bibinfo{person}{Grace Hsu}, {and} \bibinfo{person}{Amar Das}.} \bibinfo{year}{2019}\natexlab{}.
\newblock \showarticletitle{Differential privacy-enabled federated learning for sensitive health data}.
\newblock \bibinfo{journal}{\emph{arXiv preprint arXiv:1910.02578}} (\bibinfo{year}{2019}).
\newblock


\bibitem[Dwork and Roth(2014)]%
        {6-dwork2014algorithmic}
\bibfield{author}{\bibinfo{person}{Cynthia Dwork} {and} \bibinfo{person}{Aaron Roth}.} \bibinfo{year}{2014}\natexlab{}.
\newblock \showarticletitle{The algorithmic foundations of differential privacy}.
\newblock \bibinfo{journal}{\emph{Foundations and Trends{\textregistered} in Theoretical Computer Science}} \bibinfo{volume}{9}, \bibinfo{number}{3--4} (\bibinfo{year}{2014}), \bibinfo{pages}{211--407}.
\newblock


\bibitem[Jiang et~al\mbox{.}(2021)]%
        {X-jiang2021differential}
\bibfield{author}{\bibinfo{person}{Bin Jiang}, \bibinfo{person}{Jianqiang Li}, \bibinfo{person}{Guanghui Yue}, {and} \bibinfo{person}{Houbing Song}.} \bibinfo{year}{2021}\natexlab{}.
\newblock \showarticletitle{Differential privacy for industrial internet of things: Opportunities, applications, and challenges}.
\newblock \bibinfo{journal}{\emph{IEEE Internet of Things J.}} \bibinfo{volume}{8}, \bibinfo{number}{13} (\bibinfo{year}{2021}), \bibinfo{pages}{10430--10451}.
\newblock


\bibitem[Lin et~al\mbox{.}(2020)]%
        {2-lin2020differential}
\bibfield{author}{\bibinfo{person}{Ying Lin}, \bibinfo{person}{Ling-Yan Bao}, \bibinfo{person}{Ze-Minghui Li}, \bibinfo{person}{Shu-Zheng Si}, {and} \bibinfo{person}{Chao-Hsien Chu}.} \bibinfo{year}{2020}\natexlab{}.
\newblock \showarticletitle{Differential privacy protection over deep learning: An investigation of its impacted factors}.
\newblock \bibinfo{journal}{\emph{Computers \& Security}}  \bibinfo{volume}{99} (\bibinfo{year}{2020}), \bibinfo{pages}{102061}.
\newblock


\bibitem[McMahan et~al\mbox{.}(2017)]%
        {4-mcmahan2017communication}
\bibfield{author}{\bibinfo{person}{Brendan McMahan}, \bibinfo{person}{Eider Moore}, \bibinfo{person}{Daniel Ramage}, \bibinfo{person}{Seth Hampson}, {and} \bibinfo{person}{Blaise~Aguera y Arcas}.} \bibinfo{year}{2017}\natexlab{}.
\newblock \showarticletitle{Communication-efficient learning of deep networks from decentralized data}. In \bibinfo{booktitle}{\emph{Artificial intelligence and statistics}}. PMLR, \bibinfo{pages}{1273--1282}.
\newblock


\bibitem[Wei et~al\mbox{.}(2020)]%
        {8-wei2020federated}
\bibfield{author}{\bibinfo{person}{Kang Wei}, \bibinfo{person}{Jun Li}, \bibinfo{person}{Ming Ding}, \bibinfo{person}{Chuan Ma}, \bibinfo{person}{Howard~H Yang}, \bibinfo{person}{Farhad Farokhi}, \bibinfo{person}{Shi Jin}, \bibinfo{person}{Tony~QS Quek}, {and} \bibinfo{person}{H~Vincent Poor}.} \bibinfo{year}{2020}\natexlab{}.
\newblock \showarticletitle{Federated learning with differential privacy: Algorithms and performance analysis}.
\newblock \bibinfo{journal}{\emph{IEEE Trans. on Information Forensics and Security}}  \bibinfo{volume}{15} (\bibinfo{year}{2020}), \bibinfo{pages}{3454--3469}.
\newblock


\bibitem[Zhang et~al\mbox{.}(2021b)]%
        {5-zhang2021survey}
\bibfield{author}{\bibinfo{person}{Chen Zhang}, \bibinfo{person}{Yu Xie}, \bibinfo{person}{Hang Bai}, \bibinfo{person}{Bin Yu}, \bibinfo{person}{Weihong Li}, {and} \bibinfo{person}{Yuan Gao}.} \bibinfo{year}{2021}\natexlab{b}.
\newblock \showarticletitle{A survey on federated learning}.
\newblock \bibinfo{journal}{\emph{Knowledge-Based Systems}}  \bibinfo{volume}{216} (\bibinfo{year}{2021}), \bibinfo{pages}{106775}.
\newblock


\bibitem[Zhang et~al\mbox{.}(2021a)]%
        {1-zhang2021understanding}
\bibfield{author}{\bibinfo{person}{Xinwei Zhang}, \bibinfo{person}{Xiangyi Chen}, \bibinfo{person}{Mingyi Hong}, \bibinfo{person}{Zhiwei~Steven Wu}, {and} \bibinfo{person}{Jinfeng Yi}.} \bibinfo{year}{2021}\natexlab{a}.
\newblock \showarticletitle{Understanding Clipping for Federated Learning: Convergence and Client-Level Differential Privacy}.
\newblock \bibinfo{journal}{\emph{arXiv preprint arXiv:2106.13673}} (\bibinfo{year}{2021}).
\newblock


\bibitem[Zhao et~al\mbox{.}(2021)]%
        {Z-zhao2021utility}
\bibfield{author}{\bibinfo{person}{Jianzhe Zhao}, \bibinfo{person}{Keming Mao}, \bibinfo{person}{Chenxi Huang}, {and} \bibinfo{person}{Yuyang Zeng}.} \bibinfo{year}{2021}\natexlab{}.
\newblock \showarticletitle{Utility Optimization of Federated Learning with Differential Privacy}.
\newblock \bibinfo{journal}{\emph{Discrete Dynamics in Nature and Society}}  \bibinfo{volume}{2021} (\bibinfo{year}{2021}).
\newblock


\end{thebibliography}

\end{document}